\documentclass[10pt,twocolumn,letterpaper]{article}

\usepackage{iccv_cam}
\usepackage{times}
\usepackage{epsfig}
\usepackage{graphicx}
\usepackage{amsmath}
\usepackage{amssymb}
\usepackage{bm}
\usepackage{booktabs}
 \usepackage{subfigure}
\usepackage{makecell}
\usepackage{floatrow}
\usepackage{tikz}
\usepackage{array,multirow}
\usepackage{authblk}
\usepackage[accsupp]{axessibility}

\usepackage[pagebackref=true,breaklinks=true,letterpaper=true,colorlinks,bookmarks=false]{hyperref}
\renewcommand{\paragraph}[1]{\vspace{1mm}\noindent\textbf{#1}}

\iccvfinalcopy %

\def\iccvPaperID{1558} %

\ificcvfinal\fi

\begin{document}

\title{Sat2Density: Faithful Density Learning from Satellite-Ground Image Pairs}
\author{
	Ming Qian$^{1,2}$\quad  Jincheng Xiong$^{1}$\quad  Gui-Song Xia$^{1}$\quad  Nan Xue$^{2}$\thanks{Corresponding author} \vspace{-0.7em}\\
$^{1}$LIESMARS, Wuhan University\quad
$^{2}$Ant Group\\
\url{https://sat2density.github.io}
}
\maketitle

\ificcvfinal\thispagestyle{empty}\fi

\begin{abstract}
This paper aims to develop an accurate 3D geometry representation of satellite images using satellite-ground image pairs. Our focus is on the challenging problem of 3D-aware ground-views synthesis from a satellite image. We draw inspiration from the density field representation used in volumetric neural rendering and propose a new approach, called Sat2Density. Our method utilizes the properties of ground-view panoramas for the sky and non-sky regions to learn faithful density fields of 3D scenes in a geometric perspective. Unlike other methods that require extra depth information during training, our Sat2Density can automatically learn accurate and faithful 3D geometry via density representation without depth supervision. This advancement significantly improves the ground-view panorama synthesis task. Additionally, our study provides a new geometric perspective to understand the relationship between satellite and ground-view images in 3D space.

\end{abstract}

{\vspace{-0.5cm}

\section{Introduction}\label{sec:introduction}
The emergence of satellite imagery has significantly enhanced our daily lives by providing easy access to a comprehensive view of the planet. This bird's-eye view offers valuable information that compensates for the limited perspective of ground-level observations by humans. However, what specific information does satellite imagery provide, and why is it so crucial? In this paper, we propose that the most critical insights come from the analysis of the geometry, topology, and geography of cross-view observations captured by paired satellite and ground-level images. Building on this hypothesis, we aim to address the challenging problem of synthesizing ground-level images from paired satellite and ground-level imagery by leveraging density representations of 3D scenes.

\begin{figure}[ht]
    \centering
    \subfigure[Learned density from satellite images]
    {
    \centering
    \includegraphics[width=0.87\linewidth]{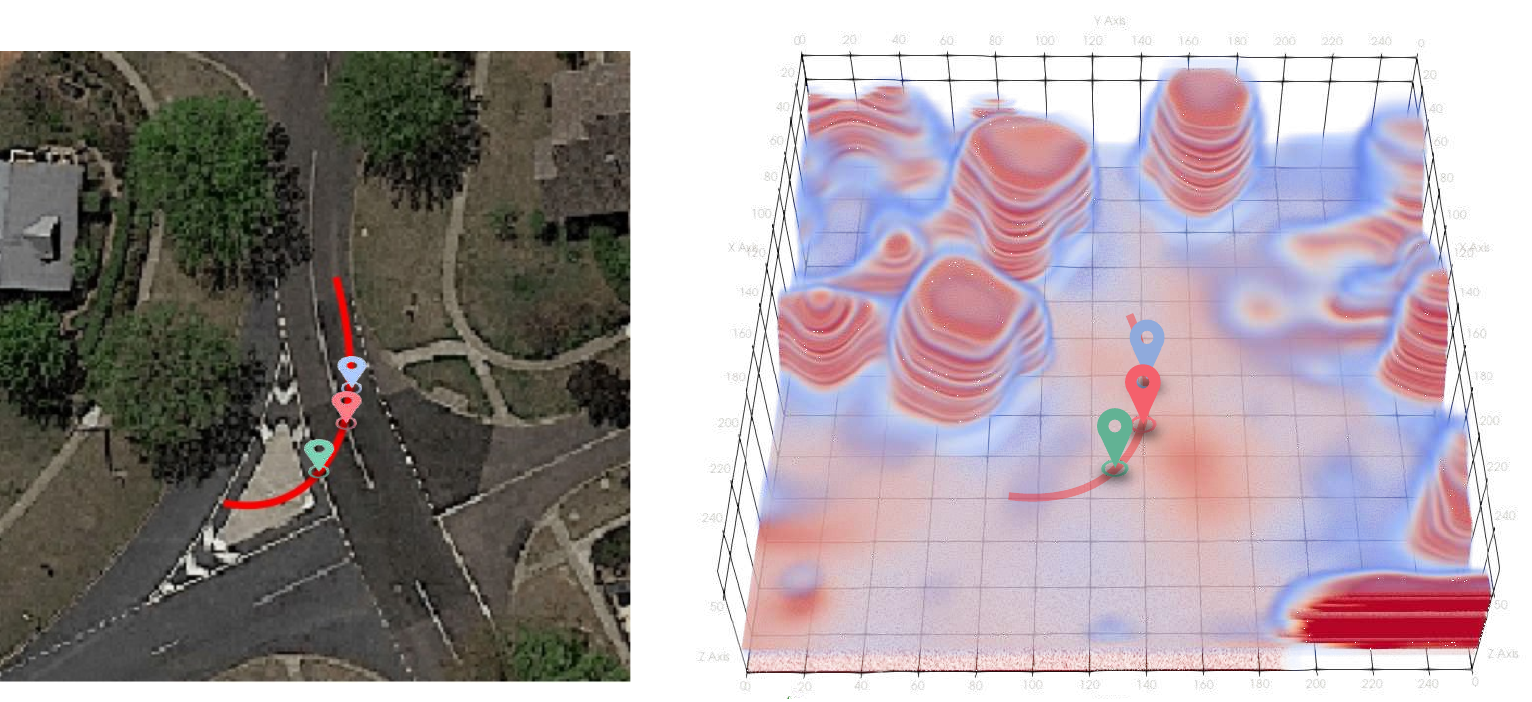}
    }
    \vspace{-0.5em}
    
    \subfigure[Synthesized panoramas]{
    \begin{minipage}[t]{0.45\linewidth}
    \includegraphics[width=1\linewidth, height = 0.5\linewidth]{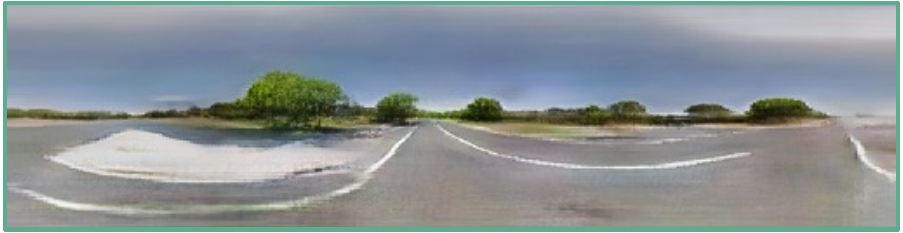}\\
    \includegraphics[width=1\linewidth, height = 0.5\linewidth]{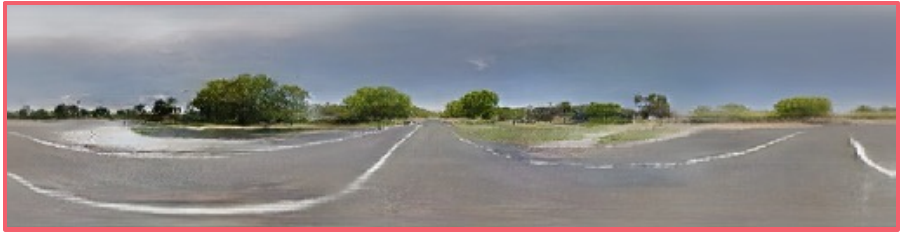}\\
    \includegraphics[width=1\linewidth, height = 0.5\linewidth]{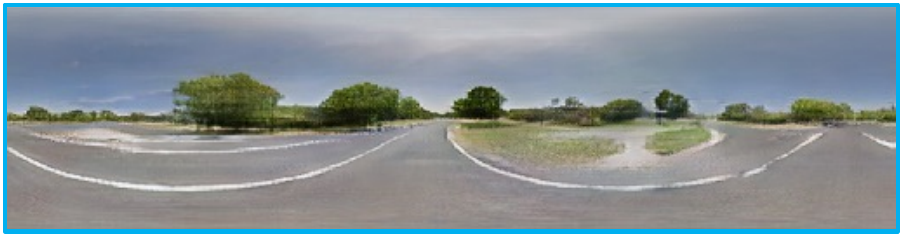}
    \end{minipage}
    }
    \subfigure[Rendered depth]{
    \begin{minipage}[t]{0.45\linewidth}
    \includegraphics[width=1\linewidth, height = 0.5\linewidth]{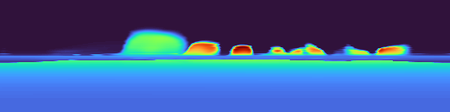}\\
    \includegraphics[width=1\linewidth, height = 0.5\linewidth]{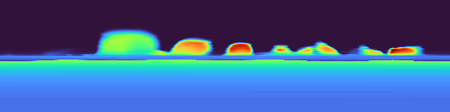}\\
    \includegraphics[width=1\linewidth, height = 0.5\linewidth]{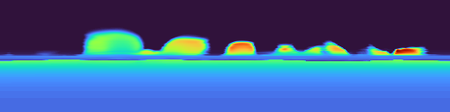}
    \end{minipage}
    }
    
    \vspace{-1em}
    
    \caption{Sat2Density trains with a collection of satellite-ground image pairs, without depth, or multi-view supervision. Our 3d GAN enables the synthesis of scenes conditioned on a satellite image, producing multi-view-consistent ground-view renderings and geometry.  
    {\em Please see the project page for more videos.}
    }
\label{fig: explain}

\end{figure}

The challenge of generating ground-level images from satellite imagery is tackled by leveraging massive datasets containing both satellite images and corresponding ground-level panoramas captured at the same geographical coordinates. However, the drastic differences in viewpoint between the two types of images, combined with the limited overlap of visual features and large appearance variations, create a highly complex and ill-posed learning problem.
To address this challenge, researchers have extensively studied the use of conditional generative adversarial networks, which leverage high-level semantics and contextual information in a generative way~\cite{Regmi_2018_CVPR, regmi_cross-view_2019,zhai2017predicting,tang2019multi, Lu_2020_CVPR}. 
However, since the contextual information used is typically at the image level, the 3D information can only be marginally inferred during training, often resulting in unsatisfactory synthesis results.

Recent studies~\cite{Shi,li2021sat2vid} have suggested that accurate 3D scene geometry plays a crucial role in generating high-quality ground-view images. With extra depth supervision, Sat2Video~\cite{li2021sat2vid} introduced a method to synthesize spatial-temporal ground-view video frames along a camera trajectory, rather than a single panorama from the center viewpoint of the satellite image. Additionally, Shi~\etal~\cite{Shi} demonstrated that coarse satellite depth maps can be learned from paired data through multi-plane image representation using a novel projection model between the satellite and ground viewpoints, but a coarse 3d representation can not facilitate rendering 3D-aware ground-view images. Building on these insights, we aim to investigate whether it is possible to achieve even more accurate 3D geometry using the vast collection of satellite-ground image pairs.

Our study is motivated by the latest developments in the neural radiance field (NeRF)~\cite{NeRF}, which has shown promising results in novel view synthesis. Benefiting from the flexibility of density field in volumetric rendering~\cite{volumetric-rendering}, faithful 3D geometry can be learned from a large number of posed images.
Therefore, we adopt density fields as the representation and focus on learning accurate density fields from paired satellite-ground image pairs.
More precisely, in this paper, we present a novel approach called Sat2Density, which involves two convolutional encode-decoder networks: DensityNet and RenderNet. The DensityNet receives satellite images as input to represent the density field in an explicit grid, which plays a crucial role in producing ground-view panorama images using the RenderNet. With such a straightforward network design, we delve into the goal of learning faithful density field first and then render high-fidelity ground-view panoramas. 

While we employed a flexible approach to representing geometry using explicit volume density and volumetric rendering, an end-to-end learning approach alone is inadequate for restoring geometry using only satellite-ground image pairs. Upon examining the tasks and satellite-ground image pairs, we identified two main factors that may impede geometry learning, which has been overlooked in previous works on satellite-to-ground view synthesis. Firstly, the sky is an essential part of ground scenes but is absent in the satellite view, and it is nearly impossible to learn a faithful representation of the infinite sky region in each image using explicit volume density. Secondly, differences in illumination among the ground images during training make it challenging to learn geometry effectively.

With the above intuitive observation, we propose two supervision signals, the \emph{ non-sky opacity supervision} and \emph{illumination injection}, to learn the density fields in a volumetric rendering form jointly. The \emph{non-sky opacity supervision} compels the density field to focus on the satellite scene and ignore the infinity regions, whereas the \emph{illumination injection} learns the illumination from sky regions to further regularize the learning density field.
By learning the density field, our Sat2Density approach goes beyond the center ground-view panorama synthesis from the training data and achieves the ground-view panorama video synthesis with the best spatial-temporal consistency. As shown in Figure~\ref{fig: explain}, our Sat2Density continuously synthesizes the panorama images along the camera trajectory. We evaluated the effectiveness of our proposed approach on two large-scale benchmarks~\cite{Shi, CVUSA} and obtained state-of-the-art performance. Comprehensive ablation studies further justified our design choices.

The main contributions of our paper are:

\begin{itemize}
\setlength{\topsep}{-1pt}
\setlength{\itemsep}{1pt}
\setlength{\parsep}{1pt}
\setlength{\parskip}{0pt}
    \item We present a geometric approach, Sat2Density, for ground-view panorama synthesis from satellite images in end-to-end learning. By explicitly modeling the challenging cross-view synthesis task in the density field for the 3D scene geometry, our Sat2Density is able to synthesize high-fidelity panoramas on camera trajectories for video synthesis without using any extra 3D information out of the training data.

    \item We tackle the challenging problem of learning high-quality 3D geometry under extremely large viewpoint changes. By analyzing the unique challenges that arise with this problem, we present two intuitive approaches \emph{non-sky opacity supervision} and \emph{illumination injection} to compel the density learning to focus on the relevant features in the satellite scene presented in the paired data while mitigating the effects of infinite regions and illumination changes.

    \item To the best of our knowledge, we are the first to successfully learn a faithful geometry representation from satellite-ground image pairs. We believe that not only do our new findings improve the performance of ground-view panorama synthesis, but the learned faithful density will also provide a renewed understanding of the relationship between satellite and ground-view image data from a 3D geometric perspective. 
    
\end{itemize}

\begin{figure*}[ht]
    \centering
    \includegraphics[width=0.95\textwidth]{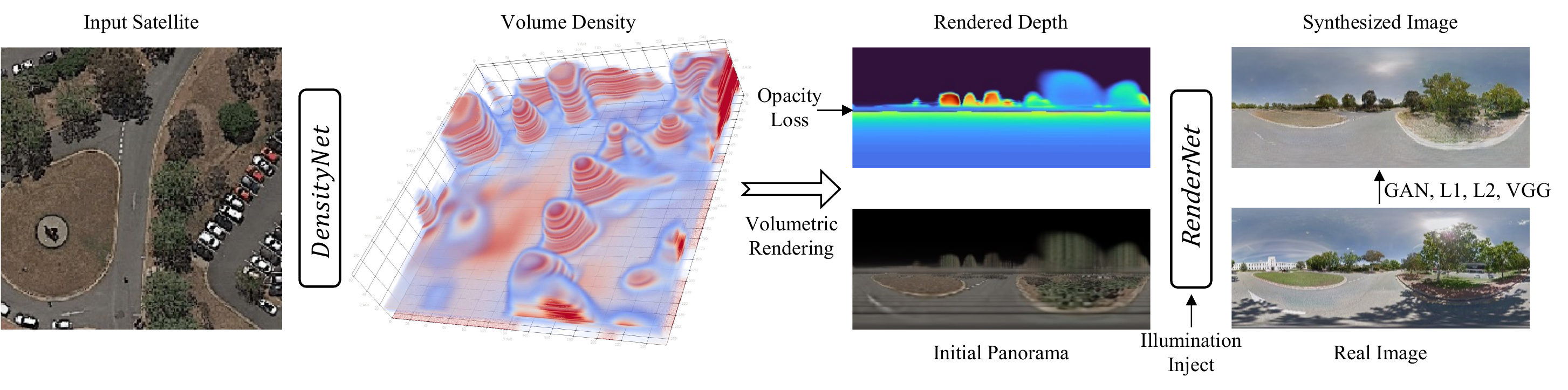}\\
    \vspace{-1mm}
    \caption{\textbf{Overview of Sat2Density.} The generation consists of two components, DensityNet and RenderNet. We optimize Sat2Density by reconstruction loss, adversarial loss, illumination injection loss, and opacity loss.
    {\em See text for details.}}
    \label{fig:Framework}
    \vspace{-1em}
\end{figure*}

\section{Related Work}
\subsection{Satellite-Ground Cross-View Perception}
Both ground-level and satellite images provide unique perspectives of the world, and their combination provided us with a more comprehensive way to understand and perceive the world from satellite-ground visual data. However, due to the drastic viewpoint changes between the satellite and ground images, poses several challenges in geo-localization~\cite{CVUSA, shi2019spatial, shi2020looking, shi2020optimal}, cross-view synthesis~\cite{Shi, li2021sat2vid, Lu_2020_CVPR, regmi_cross-view_2019, tang2019multi}, overhead image segmentation with the assistance of ground-level images~\cite{workman2022revisiting}, geo-enabled depth estimation~\cite{workman2021augmenting}, predicting ground-level scene layout from aerial imagery~\cite{zhai2017predicting}.

To address this challenge, many previous works have proposed various approaches to model and learn the drastic viewpoint changes, including the use of homography transforms~\cite{regmi_cross-view_2019}, additional depth or semantic supervision~\cite{tang2019multi, Lu_2020_CVPR,li2021sat2vid}, transformation matrices~\cite{CVUSA}, and geospatial attention~\cite{workman2022revisiting}, among others. Despite effectiveness, these approaches mainly address the challenge on the image level instead of the 3D scenes. 

Most recently, Shi \etal~\cite{Shi} proposed a method to learn geometry in satellite scenes implicitly using the height (or depth) probability distribution map, which achieved better results in synthesized road and grassland regions through their geometry projection approach. However, their learned geometry has limited effectiveness as the rendered satellite depth cannot accurately recognize objects. We go further along the line to focus on the 3D scene geometry conveyed in the satellite-ground image pairs. We demonstrate that the faithful 3D scene geometry can be explicitly decoded and leveraged with an appropriate representation and supervision signals, to obtain high-fidelity ground-view panoramas. Besides, we believe that our study brings a novel perspective to rethink satellite-ground image data for many other challenging problems.

\subsection{Neural Radiance Field}
Benefiting from the flexibility of density field in volumetric rendering~\cite{NeRF}, faithful 3D geometry can be learned from a dense number of posed images~\cite{gancraft, martin2021nerf, bautista2022gaudi, ml-gsn}. Recent works~\cite{chan2022efficient, xu2022sinnerf, yu2021pixelnerf} based on NeRF have shown that 3D representation can be learned even with only a few views. In a co-current work~\cite{wimbauer2023behind}, it is also pointed out that the flexibility of the density field helps to learn the 3D geometric structure from a single image by disentangling the color and geometry, which allows neural networks to capture reliable 3D geometry in occluded areas. %

Our goal can be viewed as an extremely challenging problem of density-based few-view synthesis with extremely large viewpoint changes, which was not studied well in previous works. In our study, we demonstrated the possibility of learning faithful geometry in the volumetric rendering formulation, shedding light on the most challenging cross-view configurations for novel view synthesis.

\section{The Proposed Sat2Density}
Figure~\ref{fig:Framework} illustrates the computation pipeline for our proposed Sat2Density. Given the input satellite image $I_{\text{sat}} \in \mathbb{R}^{H\times W\times 3}$ for the encoder-decoder DensityNet, we learn an explicit volume of the density field $V_{\sigma} \in \mathbb{R}^{H\times W\times N}$. We render the panorama depth and project the color of the satellite image along rays in the ground view to generate an initial panorama image and feed them to the RenderNet. 
To ensure consistent illumination of the synthesis, the histogram of color in the sky regions of the panorama is used as a conditional input for our method.

\subsection{Density Field Representation}
\label{volume density}
We encode an explicit volume density as a discrete representation of scene geometry and parameterize it using a plain encoder-decoder architecture in DensityNet $G_{\text{dns}}$ to learn the density field:
\begin{equation}
    V_{\sigma} = G_{\text{dns}}(I_{\text{sat}}) \ \ \ \ \ \text{s.t.}\ V_{\cdot,\cdot,\cdot}\in[0, \infty).
\end{equation}
where the density information $v = V(x,y,z)$ is stored in the volume of $V$ for the spatial location $(x,y,z)$. For any queried location that does not locate in the sample position of the explicit grid, tri-linear interpolation is used to obtain its density value. Suppose the size of the real-world cube is $(X,Y,Z)$ in the satellite image, two corner cases are considered:
1) for the locations outside the cube, we set their density to zero, and  2) we set the density in the lowest volume (\ie, V(x,y,z=0)) to a relatively large value ($10^3$ in our experiments), which made an assumption that all ground regions are solid.

With the density field representation, the volumetric rendering techniques~\cite{max1995optical} are applied to render the depth $\hat{d}$ and opacity $\hat{O}$ along the queried rays by 
\begin{equation}\label{equ: opacity}
\hat{d}=\sum_{i=1}^{S} T_{i}\alpha_{i}d_{i}, \qquad \hat{O}=\sum_{i=1}^{S} T_{i}\alpha_{i},
\end{equation} 
where $d_i$ is the distance between the camera location and the sampled position, $T_i$ denotes the accumulated transmittance along the ray at $t_i$, and $\alpha_i$ is the alpha value for the alpha compositing, written by
\begin{equation}
\alpha_{i}=1-\exp \left(-\sigma(\mathbf{x}_{i}) \delta_{i}\right) \quad T_{i}=\prod_{j=1}^{i-1}\left(1-\alpha_{j}\right).
\end{equation} 

Unlike NeRF~\cite{NeRF} that learns the radiance field to render the colored images, we take a copy-paste strategy to compute the colored images by copying the color from the satellite image along the ray via bilinear interpolation for image rendering in
\begin{equation}
    \hat{c}_{\rm map} = \sum_{i}^S T_i\alpha_i c_i,
\end{equation}
where $c_{i} = c(x_{i}, y_{i}, z_{i}) = I_{\rm sat}(\frac{x_{i}}{S_x}+\frac{H}{2}, \frac{y_{i}}{S_y}+\frac{W}{2})$. $S_x$ and $S_y$ are the scaling factors between the pixel coordinate of the satellite image and the grid coordinate in $V_{\sigma}$. 
To keep the simplicity of our Sat2Density, we did not use the hierarchical sampling along rays for the computation of depth, colors, and opacity.

Thanks to the flexibility of volumetric rendering, for the end task of ground-view panorama synthesis, it is straightforward to render the ground-view depth, opacity, and the (initial) colored image. For the subsequent RenderNet, it takes the concatenated tensor of the rendered panorama depth, opacity, and colors as input to synthesize the high-fidelity ground-view images.

Learning density could draw precise geometry information of the scene, but it is hard to acquire real density information of the satellite scene only from the satellite-ground image pairs. In our work, we propose two supervisions: {\em non-sky opacity supervision}  and {\em illumination injection} to improve the quality of the 3D geometry representation.

\subsection{Supervisions from Sky/Non-Sky Separation}
\label{Sky Opa} 
\paragraph{Non-Sky Opacity Supervision.} We draw inspiration from the study of panorama image segmentation~\cite{mihail2016sky,zhang2022bending}, which treats the sky region as a meaningful category in the segmentation task. By taking the off-the-shelf sky segmentation model~\cite{zhang2022bending} to obtain the sky masks for the training panorama images, we tackle the {\em infinity issue} with a novel {\em non-sky opacity supervision} proposed. Based on our discussion in Sec.~\ref{sec:introduction}, the pseudo sky masks provide a strong inductive basis to faithfully learn the density fields for our proposed Sat2Density in a simple way.

Denoted by $\mathcal{R}$ and $\mathcal{R}'$ the non-sky/sky regions of the ground-view panorama, the loss function $\mathcal{L}_{\rm snop}$ of our proposed non-sky opacity supervision reads to 
\begin{equation}
    \mathcal{L}_{\rm snop} = \sum_{\mathbf{r} \in \mathcal{R}}\left\|\hat{O}(\mathbf{r})-1\right\|_{1} + \sum_{\mathbf{r} \in \mathcal{R}'}\left\|\hat{O}(\mathbf{r})\right\|_{1}.
\end{equation}

\paragraph{Illumination Injection from Sky Regions. }
While the density field works well on images of static subjects captured under controlled settings, it is incapable of modeling many ubiquitous, real-world phenomena for ground-view panorama synthesis. More importantly, due to the lacking of correspondence from the sky regions in ground-view images to the paired satellite image,  we find that the variable illumination in the ground images is a key factor preventing the model to learn faithful 3D geometry.

Accordingly, we present an illumination injection from sky regions of the panorama. For the sake of simplicity of design, we choose the RGB histogram information in the sky regions as the illumination hints. In our implementation, we first cut out the sky part from the ground image, then calculate the RGB sky histogram with $N$ bins. To further exploit the representational ability of sky histograms, we follow the style encoding scheme proposed in SPADE~\cite{SPADE} to transform the sky histogram into a fixed-dim embedding, which allows our Sat2Density learn reliable information of complicated illumination from the sky histogram for the RenderNet.
From our experiments, we also find that the injection of sky histogram into the RenderNet could further improve the quality of explicit volume density by encouraging the DensityNet to focus on the stuff regions rather than being disturbed by the per-image illumination variations issue.

By combining the above two approaches, we solve the per-image illumination variations and infinity issue, and let our model focus on learning the scene geometry relationship between satellite and ground view (see Figure 
 \ref{fig:ablation study}).
Besides, thanks to the proposed sky histogram illumination injection approach,  our model could control illuminations.

\subsection{Loss Functions}
\label{loss}
Sat2Density is trained with both reconstruction loss and adversarial loss. For reconstruction loss,  we follow GAN-based syntheses works, using a combination of the perceptual loss~\cite{johnson2016perceptual}, L1, and L2 loss. 
In the adversarial loss, we use the non-saturated loss~\cite{karras2021alias} as the training objective. 
Besides, $\mathcal{L}_{snop}$ is used for opacity supervision.
For illumination learning, we follow the SPADE~\cite{park2019semantic} use a KL Divergence loss. Last, in the discriminator, we use the feature matching loss \& a modified multi-scale discriminator architecture in \cite{wang2018high}.  Details can be found in the supplemental material.

\section{Experiments}
\subsection{Implementation Details}
We train our model with $256\times256$  input satellite images and output a $256\times256\times65$ explicit volume density and finally predict a $128\times512$ 360{\textdegree} panorama image, which is the same as the setting in~\cite{Shi} for a fair comparison. The maximum height modeled by our implicit volume density is 8 meters, which is an empirical setup. We approximate the height of the street-view camera as 2 meters with respect to the ground plane, which follows Shi \etal~\cite{Shi}. The bins of histogram in each channel are 90.  The model is trained in an end-to-end manner with a batch size of 16. The optimizer we used is Adam with a learning rate of 0.00005, and $\beta_{1} = 0$, $\beta_{2} = 0.999$.
Using a 32GB Tesla V100, the training time is about 30 hours for 30 epochs. As for the architectures of DensityNet and RenderNet, they share most similarities with the networks used in Pix2Pix\cite{Pix2Pix}.
More details about the model architecture and training details can be found in the supplemental material.

\vspace{-0.3em}

\subsection{Evaluation Metrics}
\vspace{-0.2em}
We use several evaluation metrics to quantitatively assess our results. The low-level similarity measures include root-mean-square error (RMSE), structure similarity index measure (SSIM), peak signal-to-noise ratio (PSNR), and sharpness difference (SD), which evaluate the pixel-wise similarity between two images. 
We also use high-level perceptual similarity~\cite{perceptualMatric} for evaluation as in previous works. 
Perceptual similarity evaluates the feature similarity of generated and real images.
We employ the pretrained AlexNet~\cite{AlexNet} and Squeeze~\cite{SqueezeNet} networks as backbones for evaluation,
denoted as $P_{\text{alex}}$ and  $P_{\text{squeeze}}$, respectively.

\vspace{-0.3em}

\subsection{Dataset for Ground View Synthesis}
\vspace{-0.2em}
We choose CVUSA~\cite{CVUSA} and CVACT(Aligned)~\cite{Shi} datasets for comparison in the central ground-view synthesis setting, following Shi \etal~\cite{Shi}. CVACT(Aligned) is a well-posed dataset aligned in Shi \etal~\cite{Shi}, with 360{\textdegree} horizontal and 180{\textdegree} vertical visualization in panorama ground truth. Hence, we selected it for controllable illumination visualization and controllable video generation. 
For the dataset CVUSA, we only use it for center-ground view synthesis as their upper and lower parts of the panoramic image are trimmed for the geo-localization task, and the number of trimmed pixels is unknown~\cite{CVUSA}.
During training and testing, we considered the street-view panoramas in the CVUSA dataset as having a 90{\textdegree} vertical field of view (FoV), with the central horizontal line representing the horizon. CVACT(Aligned) contains 26,519 training pairs and 6,288 testing pairs, while CVUSA contains 35,532 training pairs and 8,884 testing pairs. We did not choose other available datasets built in the city scene, such as OP\cite{OP} and VIGOR\cite{zhu2021vigor}, since their GPS information is less accurate in urban areas compared to open rural areas, and poorly posed satellite-ground image pairs are not suitable for our task.%

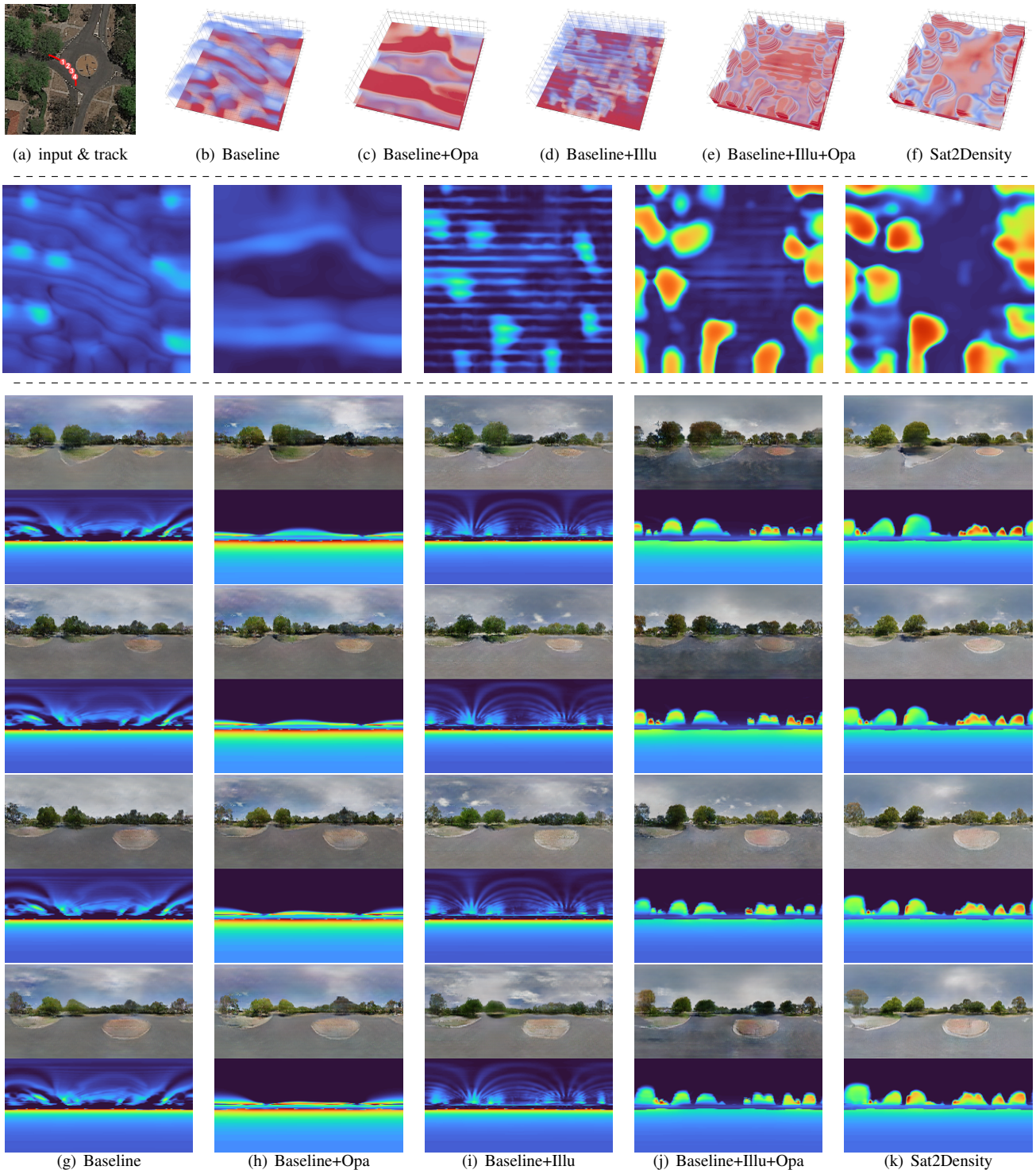
\begin{figure*}[ht]
    \vspace{-1em}
    \centering
    
    \subfigure[input \& track]{
    \captionsetup{font={footnotesize}}
    \centering
    \includegraphics[width=0.125\linewidth, height = 0.125\linewidth]{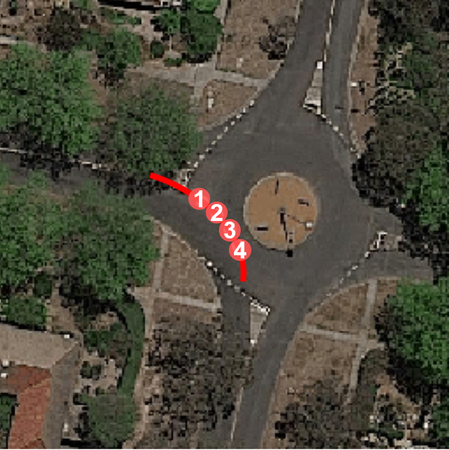}
    }
    \hfill
    \subfigure[Baseline]{
    \centering
    \includegraphics[width=0.15\linewidth, height = 0.125\linewidth]{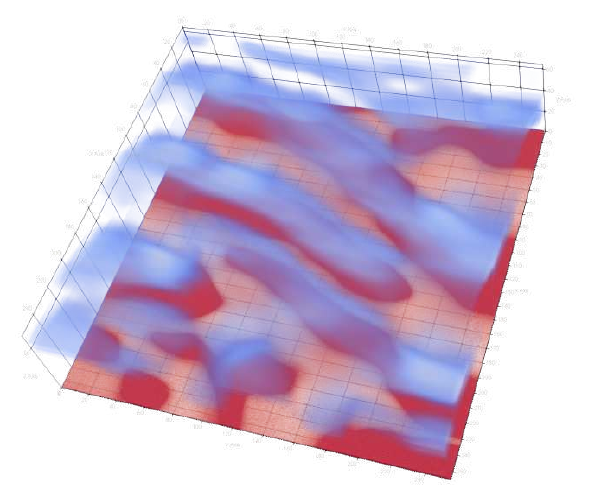}
    }
    \hfill
    \subfigure[Baseline+Opa]{
    \centering
    \includegraphics[width=0.15\linewidth, height = 0.125\linewidth]{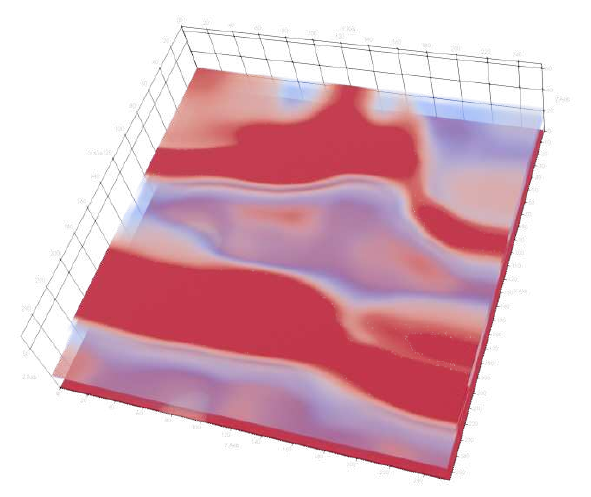}
    }
    \hfill
    \subfigure[Baseline+Illu]{
    \centering
    \includegraphics[width=0.15\linewidth, height = 0.125\linewidth]{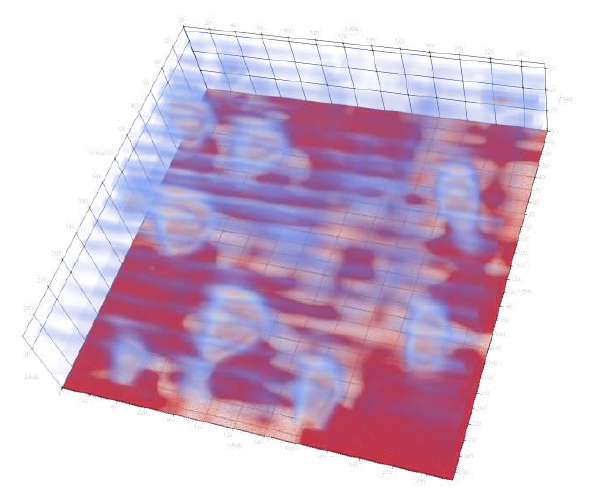}
    }
    \hfill
    \subfigure[Baseline+Illu+Opa]{
    \centering
    \includegraphics[width=0.15\linewidth, height = 0.125\linewidth]{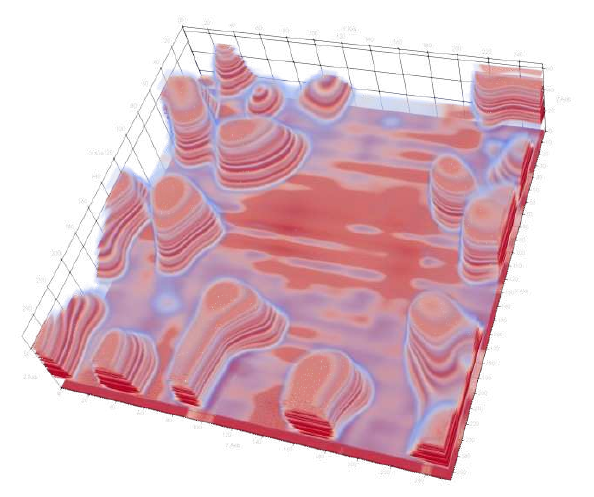}
    }
    \hfill
    \subfigure[Sat2Density]{
    \centering
    \includegraphics[width=0.15\linewidth, height = 0.125\linewidth]{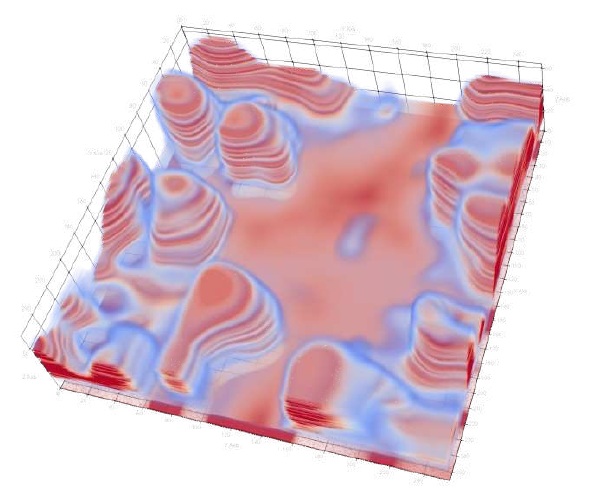}
    }

    \begin{tikzpicture}
    \draw[dashed] (0,0) -- (17,0);
    \end{tikzpicture}

    \vspace{0.3em}

    \hspace{-0.55em}
    \includegraphics[width=0.18\linewidth]{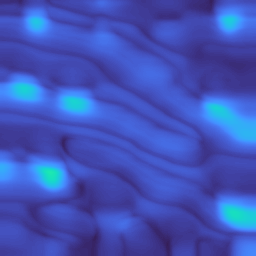}
    \hspace{0.6em}
    \includegraphics[width=0.18\linewidth]{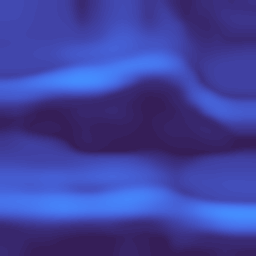}
    \hspace{0.6em}
    \includegraphics[width=0.18\linewidth]{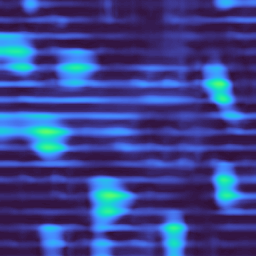}
    \hspace{0.6em}
    \includegraphics[width=0.18\linewidth]{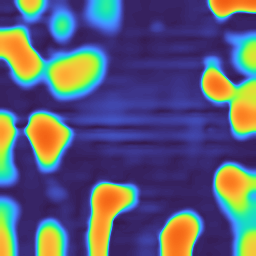}
    \hspace{0.6em}
    \includegraphics[width=0.18\linewidth]{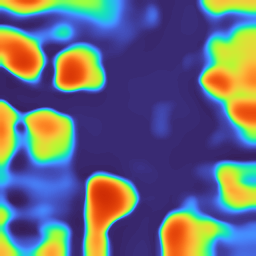}
    
    \vspace{-0.7em}

    \begin{tikzpicture}
    \draw[dashed] (0,0) -- (17,0);
    \end{tikzpicture}

    \subfigure[Baseline]{
    \begin{minipage}[t]{0.18\linewidth}
    \includegraphics[width=1\linewidth, height = \linewidth]{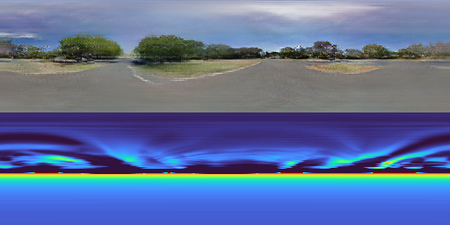}\\
    \includegraphics[width=1\linewidth, height = \linewidth]{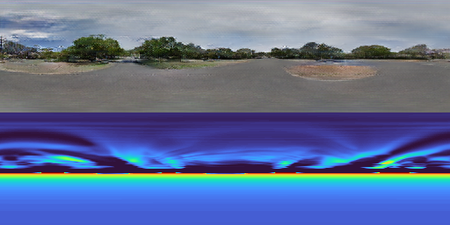}\\
    \includegraphics[width=1\linewidth, height = \linewidth]{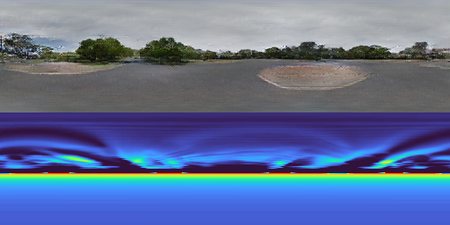}\\
    \includegraphics[width=1\linewidth, height = \linewidth]{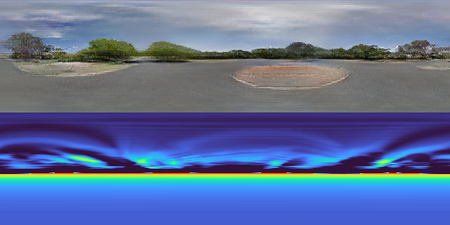}
    \end{minipage}
    }
    \hfill
    \subfigure[Baseline+Opa]{
    \begin{minipage}[t]{0.18\linewidth}
    \includegraphics[width=1\linewidth, height = \linewidth]{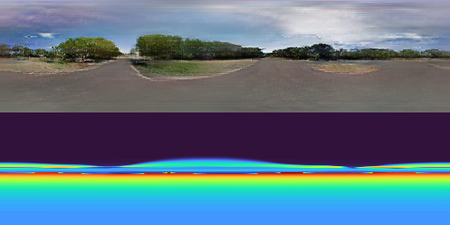}\\
    \includegraphics[width=1\linewidth, height = \linewidth]{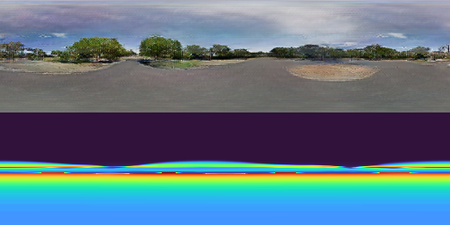}\\
    \includegraphics[width=1\linewidth, height = \linewidth]{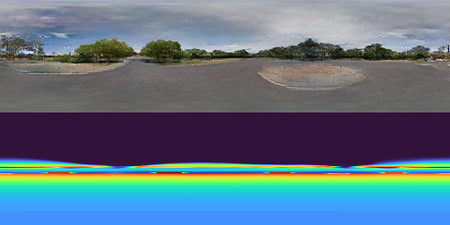}\\
    \includegraphics[width=1\linewidth, height = \linewidth]{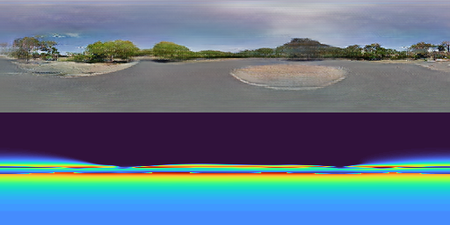}
    \end{minipage}
    }
    \hfill
    \subfigure[Baseline+Illu]{
    \begin{minipage}[t]{0.18\linewidth}
    \includegraphics[width=1\linewidth, height = \linewidth]{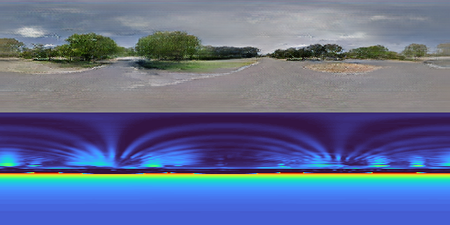}\\
    \includegraphics[width=1\linewidth, height = \linewidth]{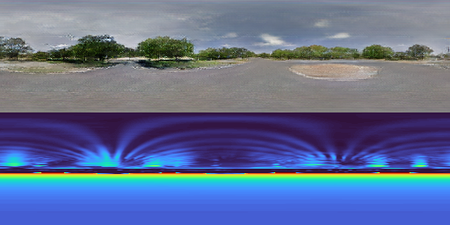}\\
    \includegraphics[width=1\linewidth, height = \linewidth]{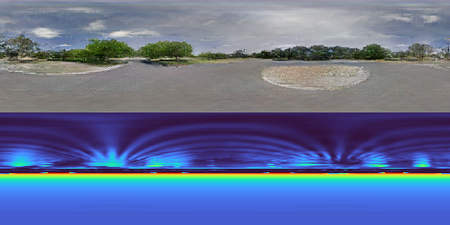}\\
    \includegraphics[width=1\linewidth, height = \linewidth]{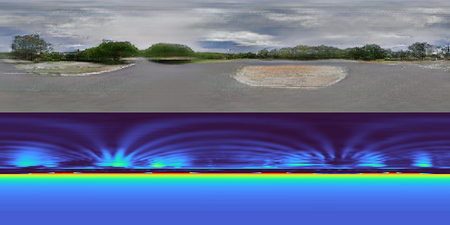}
    \end{minipage}
    }
    \hfill
    \subfigure[Baseline+Illu+Opa]{
    \begin{minipage}[t]{0.18\linewidth}
    \includegraphics[width=1\linewidth, height = \linewidth]{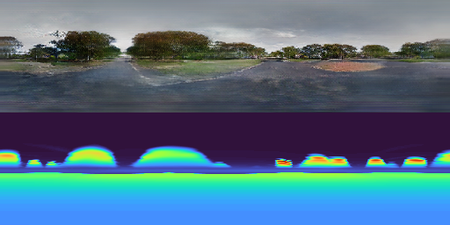}\\
    \includegraphics[width=1\linewidth, height = \linewidth]{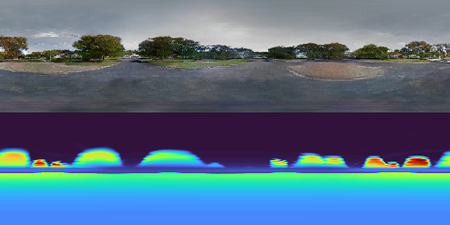}\\
    \includegraphics[width=1\linewidth, height = \linewidth]{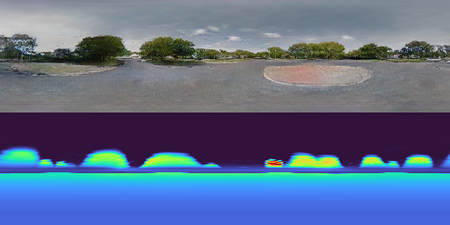}\\
    \includegraphics[width=1\linewidth, height = \linewidth]{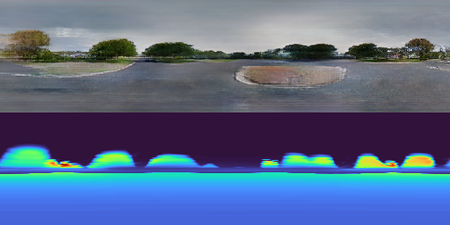}
    \end{minipage}
    }
    \hfill
    \subfigure[Sat2Density]{
    \begin{minipage}[t]{0.18\linewidth}
    \includegraphics[width=1\linewidth, height = \linewidth]{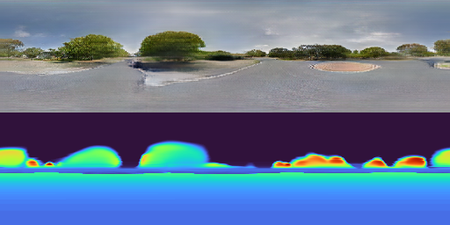}\\
    \includegraphics[width=1\linewidth, height = \linewidth]{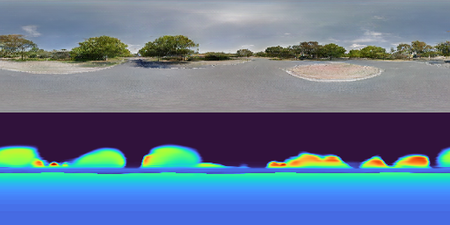}\\
    \includegraphics[width=1\linewidth, height = \linewidth]{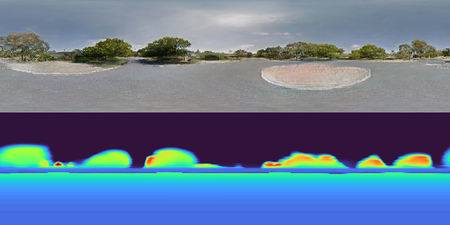}\\
    \includegraphics[width=1\linewidth, height = \linewidth]{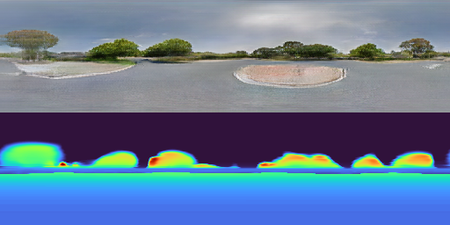}
    \end{minipage}
    }
    \vspace{-1em}
    \caption{Ablation study on CVACT (Aligned) dataset. In the first row, the picture on the upper left is the input image. Each point from left to right is related to the bottom four rows from up to down. The remaining five images in the first row are the density rendered from the input satellite image following the setting (b-f) one by one.
    The images in the second row are the satellite depth calculated following the setting (b-f) one by one. `Baseline' means baseline, `Opa' means add {\em non-sky opacity supervision}, `Illu' means add {\em illumination injection}, and `Sat2Density' is our final result, compared to `Baseline+Illu+Opa', we concatenate the depth map and initial panorama together to send to the RenderNet rather than only the initial panorama. {\em The video could be seen in the project page.}}
    \label{fig:ablation study}
\end{figure*}

\begin{figure*}[t!]
    \centering
    \subfigure[$I_{\text{sat}}$]{
    \begin{minipage}[t]{0.08\linewidth}
    \includegraphics[width=1\linewidth]{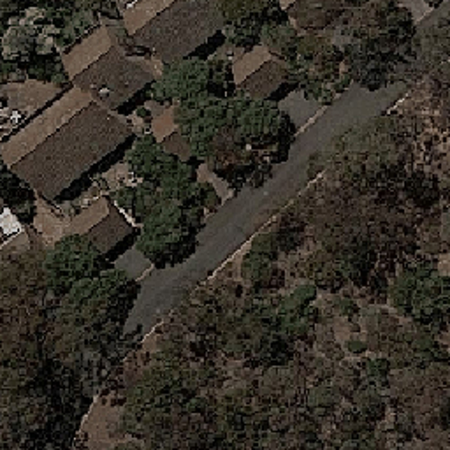}\\
    \includegraphics[width=1\linewidth]{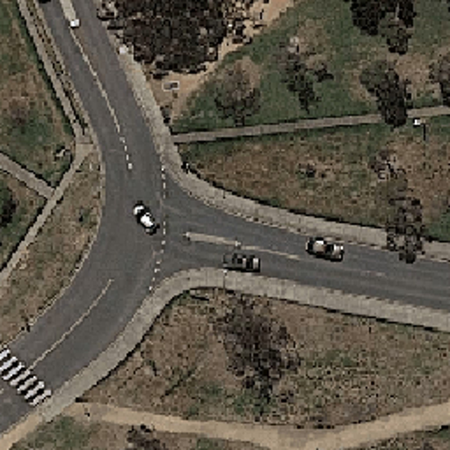}\\
    \includegraphics[width=1\linewidth]{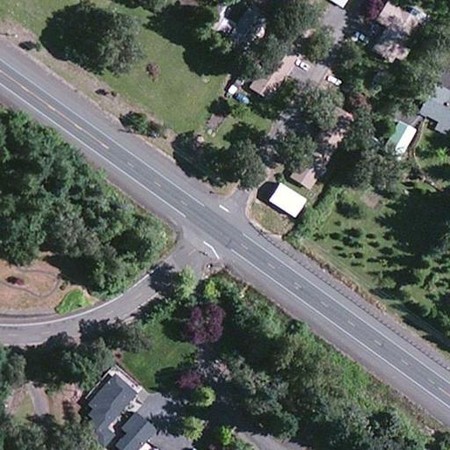}\\
    \includegraphics[width=1\linewidth]{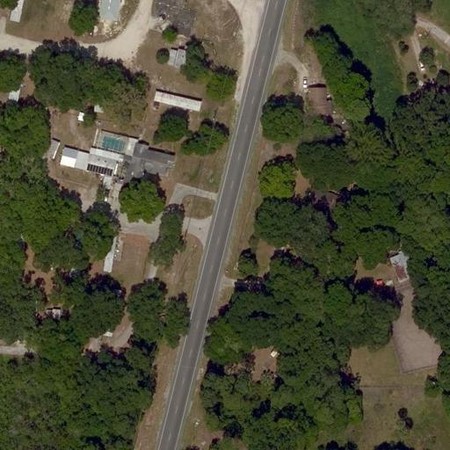}
    \end{minipage}
    }
    \subfigure[Pix2Pix~\cite{Pix2Pix}]{
    \begin{minipage}[t]{0.16\linewidth}
    \includegraphics[width=1\linewidth, height = 0.5\linewidth]{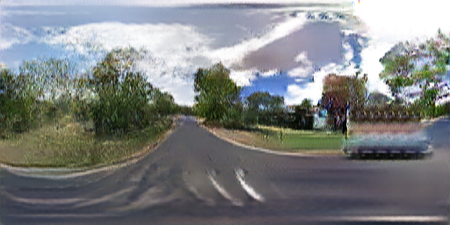}\\
    \includegraphics[width=1\linewidth, height = 0.5\linewidth]{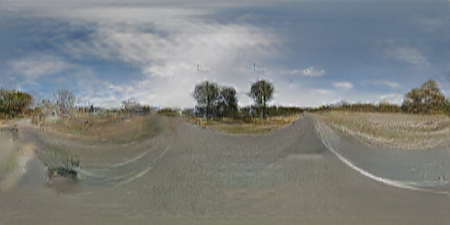}\\
    \includegraphics[width=1\linewidth, height = 0.5\linewidth]{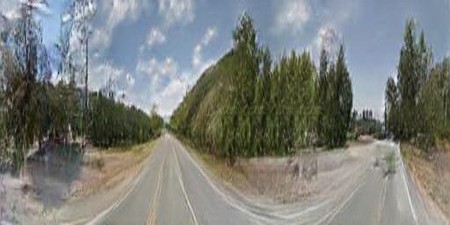}\\
    \includegraphics[width=1\linewidth, height = 0.5\linewidth]{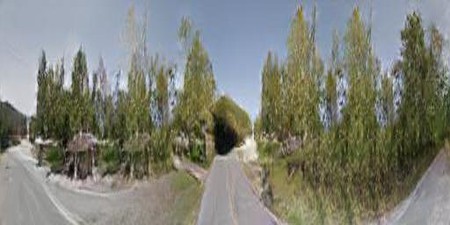}
    \end{minipage}
    }
    \subfigure[XFork~\cite{Regmi_2018_CVPR}]{
    \begin{minipage}[t]{0.16\linewidth}
    \includegraphics[width=1\linewidth, height = 0.5\linewidth]{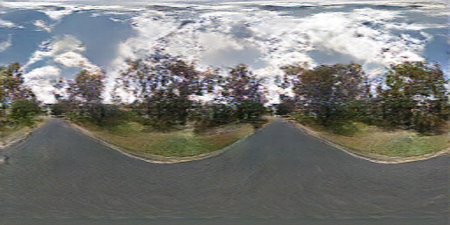}\\
    \includegraphics[width=1\linewidth, height = 0.5\linewidth]{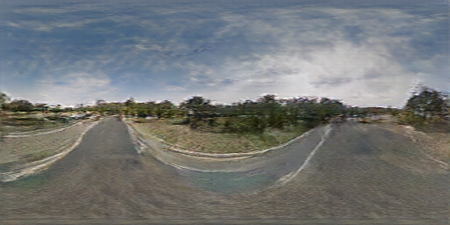}\\
    \includegraphics[width=1\linewidth, height = 0.5\linewidth]{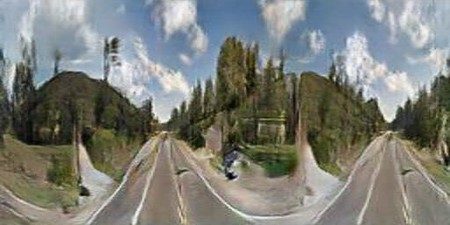}\\
    \includegraphics[width=1\linewidth, height = 0.5\linewidth]{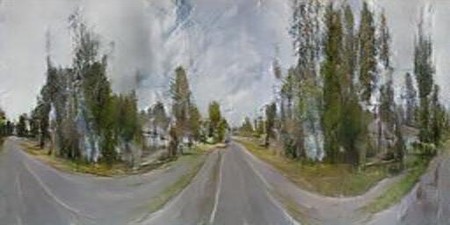}
    \end{minipage}
    }
    \subfigure[Shi \etal~\cite{Shi}]{
    \begin{minipage}[t]{0.16\linewidth}
    \includegraphics[width=1\linewidth, height = 0.5\linewidth]{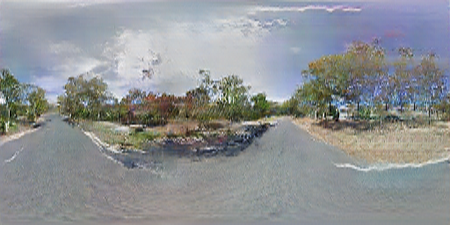}\\
    \includegraphics[width=1\linewidth, height = 0.5\linewidth]{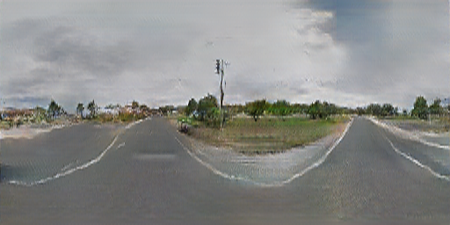}\\
    \includegraphics[width=1\linewidth, height = 0.5\linewidth]{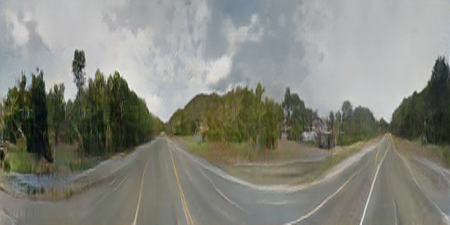}\\
    \includegraphics[width=1\linewidth, height = 0.5\linewidth]{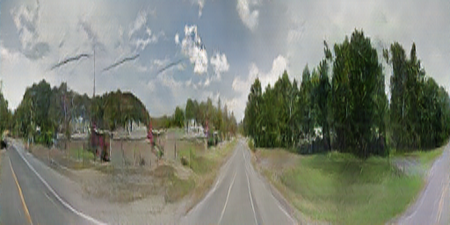}
    \end{minipage}
    }
    \subfigure[Sat2Density]{
    \begin{minipage}[t]{0.16\linewidth}
    \includegraphics[width=1\linewidth, height = 0.5\linewidth]{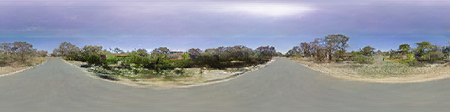}\\
    \includegraphics[width=1\linewidth, height = 0.5\linewidth]{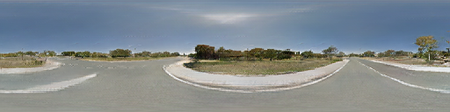}\\
    \includegraphics[width=1\linewidth, height = 0.5\linewidth]{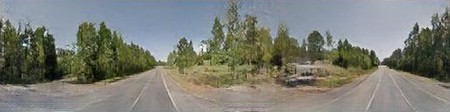}\\
    \includegraphics[width=1\linewidth, height = 0.5\linewidth]{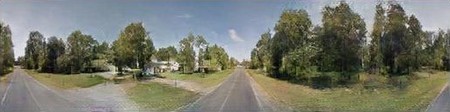}
    \end{minipage}
    }
    \subfigure[Ground Truth]{
    \begin{minipage}[t]{0.16\linewidth}
    \includegraphics[width=1\linewidth, height = 0.5\linewidth]{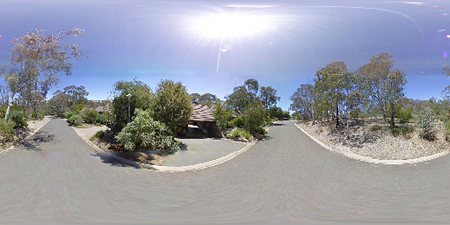}\\
    \includegraphics[width=1\linewidth, height = 0.5\linewidth]{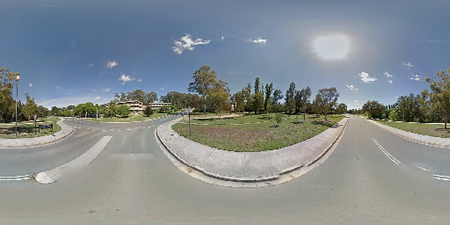}\\
    \includegraphics[width=1\linewidth, height = 0.5\linewidth]{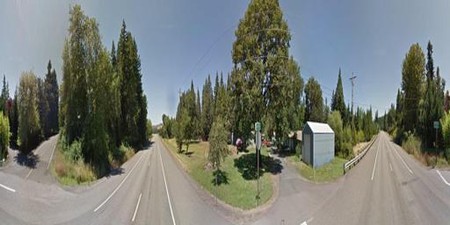}\\
    \includegraphics[width=1\linewidth, height = 0.5\linewidth]{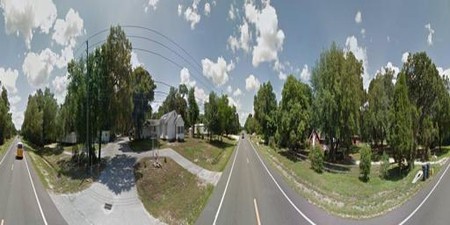}
    \end{minipage}
    }
    \caption{Example images generated by different methods in center panorama synthesis task. The top two rows show the results on CVACT (Aligned) dataset, and the bottom two rows show the results on the CVUSA dataset.}
    \label{fig:demo}
    \vspace{-1em}
\end{figure*}

\vspace{-0.3em}
\subsection{Ablation Study}
\vspace{-0.2em}

In this section, we conduct experiments to validate the importance of each component in our framework, including {\em non-sky opacity supervision}, {\em illumination injection}, and whether to concatenate depth with the initial panorama before sending it to the RenderNet. We first present quantitative comparisons in Table \ref{tab:ablation study} for the center ground-view synthesis setting. It is evident that the illumination injection most affects  the quantitative result, at the same time, only adding non-sky opacity supervision will lead to a little drop in the quantitative score. But combining the two approaches will lead to better scores. 
Moreover, the comparison on whether concatenate depth to the RenderNet shows almost equal results in terms of quantitative comparison.

Figure \ref{fig:ablation study} shows some samples from the rendered panorama video and their corresponding depth maps. The results show that without the proposed components (baseline), the rendered depth seems meaningless in the upper half, while the lower regions look good. We attribute this phenomenon to the fact that the lower bound of the panorama is the ray that looks down, which is highly related to the ground region near the shooting center in the satellite. It can be easily learned by a simple CNN with a simple geometry projection, which also explains why the work in~\cite{Shi} can render the ground region well.

Compared to the baseline, adding the illumination injection can make the rendered depth look better, but the trees' density looks indistinct, and the sky region's density is still unclear.
While only adding non-sky opacity supervision, the air region's opacity turns to zero, but the area between air and the ground is still barely satisfactory. The supervision did clear the sky region in the volume density, but the inner region between the sky and the ground is also smoothed. This is because  such coarse supervision cannot help the model recognize the complex region.

By combining both strategies (Baseline+Illu+Opa), we can achieve a plausible 3D geometry representation that can generate a depth map faithful to the satellite image and the reference illumination. The volume density is clear compared to the above settings, and we can easily distinguish the inner regions.

Furthermore, when depth is incorporated into the rendering process, the resulting images tend to emphasize regions with depth information. This reduces the likelihood of generating objects in infinite areas randomly and leads to a synthesized ground view that more closely resembles the satellite scene, which can be observed from the video.

\begin{table}[t!]
\centering
\vspace{-2mm}
\resizebox{0.99\linewidth}{!}{
\begin{tabular}{lcccccc}
\hline
      Comparison  & RMSE $\downarrow$ & SSIM$\uparrow$ & PSNR$\uparrow$ & SD $\uparrow$ & $P_{alex}\downarrow$ &  $P_{squ}\downarrow$ \\ 
      \hline
    Base          & 48.40 & 0.4491& 14.67& 12.76& 0.3772& 0.2486 \\
    Base+Opa       & 48.39 & 0.4431& 14.63& 12.70& 0.3847& 0.2525 \\
    Base+Illu       & 41.62 & 0.4689& 15.96& 12.90& 0.3497& 0.2225 \\
    Base+Opa+Illu    & 40.71 & 0.4710& 16.16& 12.83& \textbf{0.3329}& 0.2154 \\
    Sat2Density & \textbf{39.76} & \textbf{0.4818}& \textbf{16.38}& \textbf{12.90}& 0.3339& \textbf{0.2145} \\\hline
\end{tabular}}
\caption{Ablation study results on CVACT (Aligned) dataset. `Base' means baseline, `Opa' means add {\em non-sky opacity supervision}, `Illu' means add {\em illumination injection}, and `Sat2Density' is our result, compared to `Baseline+Opa+Illu', we concatenate the depth map and initial panorama together to send to the RenderNet rather than only the initial panorama.}
\label{tab:ablation study}
\vspace{-1.5mm}
\end{table}

\vspace{-0.3em}
\subsection{Center Ground-View Synthesis Comparison}
\vspace{-0.2em}
In the center ground-view synthesis setting, we compare our method with Pix2Pix~\cite{Pix2Pix}, XFork~\cite{Regmi_2018_CVPR}, and Shi \etal~\cite{Shi}. Pix2Pix and XFork are classic GAN-based models for image-to-image translation, but ignore the 3D geometry connections between the two views. Shi \etal~\cite{Shi} is the first geometry-guided synthesis model, which represents the 3D geometry in the depth probability MPI, showing brilliant results in the center ground-view synthesis setting.

\begin{table}[t!]
\centering
\vspace{-2mm}
\resizebox{0.99\linewidth}{!}{
\begin{tabular}{c|cccccccc}
\hline
       &Method      & RMSE $\downarrow$ & SSIM$\uparrow$ & PSNR$\uparrow$ & SD $\uparrow$ & $P_{\text{alex}}\downarrow$ &  $P_{\text{squeeze}}\downarrow$ &\thead{Inference\\time/ms} \\  \hline

        \parbox[t]{2mm}{\multirow{5}{*}{\rotatebox[origin=c]{90}{CVACT (Ali.)}}} & Pix2Pix~\cite{Pix2Pix}  & 49.75 & 0.3852 & 14.38 & 12.09 & 0.4654 & 0.3096 & 10.29\\
     & XFork~\cite{Regmi_2018_CVPR}  & 48.95 & 0.3710 & 14.50 &  12.32 & 0.4638 & 0.3262 & 17.24\\
     & Shi \etal~\cite{Shi} & 48.50 & 0.4272 & 14.59 & 12.31   & 0.4059 & 0.2708 & 109.88\\    \cline{2-9}  
     & Sat2Density &47.13  & 0.4586 & 14.92 & 12.77   & 0.3842 & 0.2573 & - \\ 
      & Sat2Density-oracle    & \textbf{39.76} & \textbf{0.4818} & \textbf{16.38} & \textbf{12.90}   & \textbf{0.3339} & \textbf{0.2145} & 33.12\\  \hline

 \parbox[t]{2mm}{\multirow{5}{*}{\rotatebox[origin=c]{90}{CVUSA}}} & Pix2Pix     & 55.27 & 0.2946 & 13.48 & 11.97 & 0.5092 & 0.3902 &-\\
      & XFork       & 54.11 & 0.2873 & 13.68 & 12.15 & 0.5144 & 0.4041 & - \\
      & Shi \etal     & 53.75 & 0.3451 & 13.75 & 12.06 & 0.4639 & 0.3506  & - \\  \cline{2-9}
      & Sat2Density   & 53.30 & 0.3301 & 13.78 & 12.38 & 0.4504 & 0.3365  & - \\   
      & Sat2Density-oracle   & \textbf{48.75} & \textbf{0.3584} & \textbf{14.66} & \textbf{12.53}   & \textbf{0.4163}    & \textbf{0.3058} & -  \\ \hline
\end{tabular}
}
\caption{Quantitative comparison with existing algorithms on the CVACT (Aligned) and CVUSA datasets in center ground-view synthesis setting. `Sat2Density'  means we randomly choose nine histograms from the test set when inference and then average it. `Sat2Density-oracle' means the histogram is from the GT ground image. The inference time was tested on a Tesla V100.}
\label{tab:result}
\vspace{-1.5mm}
\end{table}

\begin{figure*}[t!]
    \centering
    \includegraphics[width=0.100\linewidth, height = 0.100\linewidth]{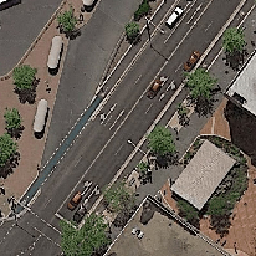}
    \includegraphics[width=0.17\linewidth, height = 0.100\linewidth]{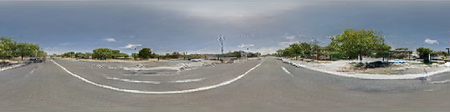}
    \includegraphics[width=0.17\linewidth, height = 0.100\linewidth]{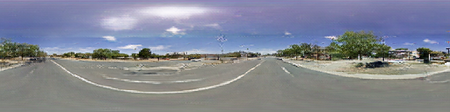}
    \includegraphics[width=0.17\linewidth, height = 0.100\linewidth]{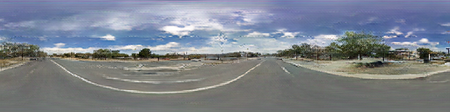}
    \includegraphics[width=0.17\linewidth, height = 0.100\linewidth]{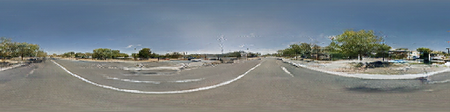}
    \includegraphics[width=0.17\linewidth, height = 0.100\linewidth]{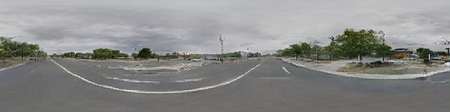}
    
    \includegraphics[width=0.100\linewidth, height = 0.100\linewidth]{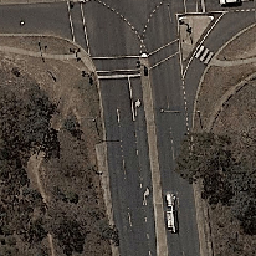}
    \includegraphics[width=0.17\linewidth, height = 0.100\linewidth]{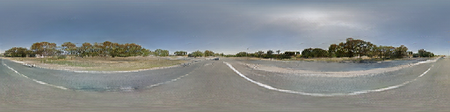}
    \includegraphics[width=0.17\linewidth, height = 0.100\linewidth]{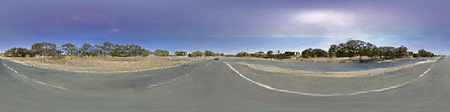}
    \includegraphics[width=0.17\linewidth, height = 0.100\linewidth]{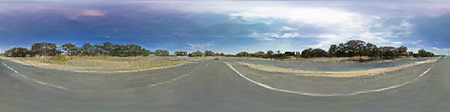}
    \includegraphics[width=0.17\linewidth, height = 0.100\linewidth]{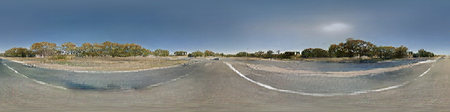}
    \includegraphics[width=0.17\linewidth, height = 0.100\linewidth]{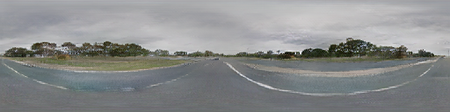}

    \caption{{Controllable illumination synthesis}: each  row shows the results rendered from the same satellite image, and each column shares the same illumination from the same ground image.}
    \label{fig: style synthesis}
    \vspace{-0.5em}
\end{figure*}

\paragraph{Quantitative Comparison.}
As presented in Table \ref{tab:result}, it is evident that Sat2Density achieves the best performance on all scores, including both low-level and perceptual similarity measures. Even choosing illumination hint randomly, our model still outperforms other methods. 

Moreover, a combined analysis of the quantitative results of Sat2Density-sin and controllable illumination in Figure~\ref{fig: style synthesis} reveals that illumination can significantly affect both common low-level and perceptual similarity measures, although the objects in the scene remain unchanged. As a result, it is more important to consider qualitative comparisons and video synthesis results.

\paragraph{Qualitative Comparison.}
In Figure \ref{fig:demo},
we find that the condition-GAN based methods can only synthesize good-looking ground images, but can not restore the geometry information from the satellite scene. Shi ~\etal\cite{Shi} learn a coarse geometry representation, so the 3D information in the ground region is more reliable. %
For our method, as discussed in the ablation study, the high-fidelity synthesis (especially in the most challenging regions between the sky and the ground) is approached by learning faithful density representation of the 3D space.

\vspace{-0.3em}
\subsection{Controllable Illumination}
\vspace{-0.2em}
As shown in Figure \ref{fig: style synthesis}, the sky histogram could easily control the image's illumination, while the semantics did not change, \eg The road's color was changed by giving different illumination, but the shape remains unchanged. Besides, the illumination interpolation results can be seen on the project page, which also show the superiority of illumination injection.

\vspace{-0.3em}
\subsection{Ground Video Generation}
\vspace{-0.2em}
In Figure \ref{fig: depth compare}, we compare the rendered satellite depth, and synthesized ground images from a camera trajectory with the expansion of Shi \etal~\cite{Shi}. Shi \etal~\cite{Shi} focus on synthesizing ground panorama in the center of the satellite image, as they learn geometry by depth probability map, we expand their work by moving the camera position when inference.  We also show the rendered depth maps. It is worth noting that Shi \etal~\cite{Shi} cannot generate a depth map for novel views, due to the intrinsic flaw of the depth probability representation. 

\begin{figure}[t]
    \vspace{-1em}
    \centering
    
    \subfigure[input image \& track]{
    \captionsetup{font={footnotesize}}
    \centering
    \includegraphics[width=0.3\linewidth, height = 0.3\linewidth]{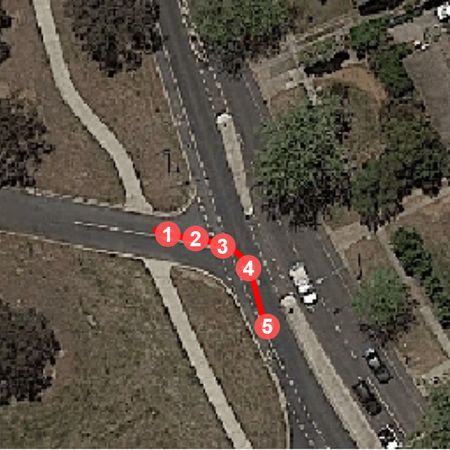}
    }
    \subfigure[Ours sat depth]{
    \centering
    \includegraphics[width=0.3\linewidth, height = 0.3\linewidth]{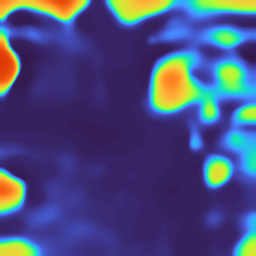}
    }
    \subfigure[Shi \etal sat depth]{
    \centering
    \includegraphics[width=0.3\linewidth, height = 0.3\linewidth]{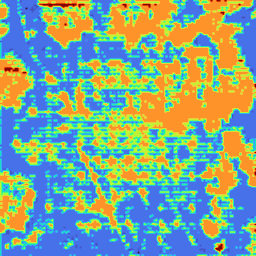}
    }
    \vspace{-1em}

    \subfigure[Our ground depth]{
    \begin{minipage}[t]{0.30\linewidth}
    \includegraphics[width=1\linewidth, height = 0.5\linewidth]{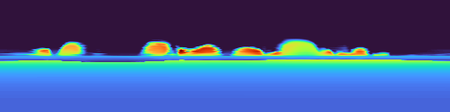}\\
    \includegraphics[width=1\linewidth, height = 0.5\linewidth]{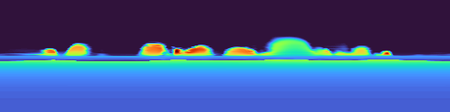}\\
    \includegraphics[width=1\linewidth, height = 0.5\linewidth]{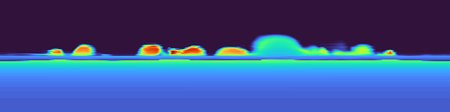}\\
    \includegraphics[width=1\linewidth, height = 0.5\linewidth]{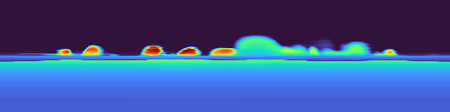}\\
    \includegraphics[width=1\linewidth, height = 0.5\linewidth]{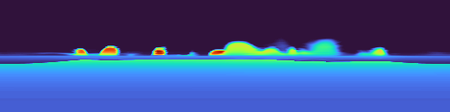}
    \end{minipage}
    }
    \subfigure[Our ground]{
    \begin{minipage}[t]{0.30\linewidth}
    \includegraphics[width=1\linewidth, height = 0.5\linewidth]{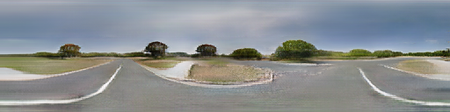}\\
    \includegraphics[width=1\linewidth, height = 0.5\linewidth]{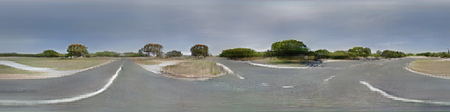}\\
    \includegraphics[width=1\linewidth, height = 0.5\linewidth]{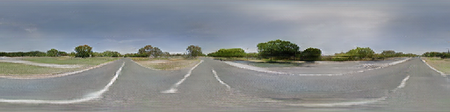}\\
    \includegraphics[width=1\linewidth, height = 0.5\linewidth]{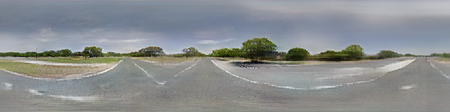}\\
    \includegraphics[width=1\linewidth, height = 0.5\linewidth]{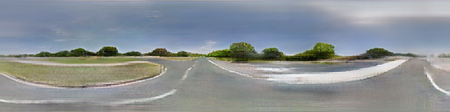}
    \end{minipage}
    }
    \subfigure[Shi \etal]{
    \begin{minipage}[t]{0.30\linewidth}
    \includegraphics[width=1\linewidth, height = 0.5\linewidth]{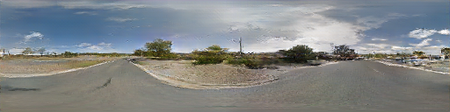}\\
    \includegraphics[width=1\linewidth, height = 0.5\linewidth]{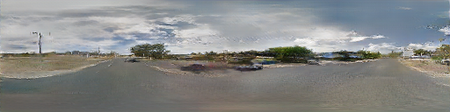}\\
    \includegraphics[width=1\linewidth, height = 0.5\linewidth]{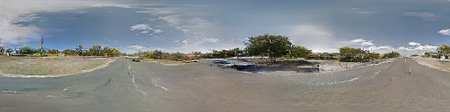}\\
    \includegraphics[width=1\linewidth, height = 0.5\linewidth]{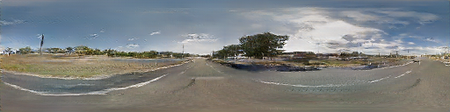}\\
    \includegraphics[width=1\linewidth, height = 0.5\linewidth]{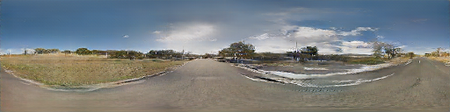}
    \end{minipage}
    }
\vspace{-1em}
\caption{Synthesized video \& depth comparison. (a) is the input satellite image, the red curve is the camera trajectory to synthesize video, the red point is chosen for visualization in (d-f), (b-c) is the rendered satellite depth by our method and Shi \etal, (d) is the rendered depth of the ground image by our method, (e), (f) are the rendered ground image by our method and Shi \etal separately. {\em The video can be seen in the project page.}
}
\label{fig: depth compare}
\end{figure}

From the synthesized satellite depth, we observe that Shi \etal~\cite{Shi} can only render a very coarse satellite depth, and is hard to recognize most regions. In contrast, trees and ground regions can easily distinguish from our satellite depth, and the depth in ground regions appears smooth. Additionally, we can render depth in any view direction by volume rendering, as shown in Figure \ref{fig: depth compare} (d).

Furthermore, we find that the rendered ground video by the expanded Shi \etal~\cite{Shi} has little consistency due to the unfaithful 3D geometry representation, as evidenced by the inconsistencies present in the trees and sky. These results demonstrate that Sat2Density is capable of rendering temporal and spatially consistent videos.

\vspace{-2mm}
\section{Discussion and Limitations}
\vspace{-2mm}
Although Sat2Density is advancing the state-of-the-art, it still has some limitations. For instance, the tree density and visibility of houses are not perfect in our results, which would come down to the following reasons. Firstly, the one-to-one satellite and ground image pairs are not optimal for training, as having multiple ground images corresponding to one satellite image would be better. Additionally, images taken on different days may introduce transient objects that our approach is unable to handle. 
Then, the projected color map sent to the RenderNet may be too coarse in the region between sky and ground, which could impact the final result. 
Finally, well-aligned image pairs are required to learn geometry, so we are unable to evaluate the effectiveness of our approach in city scenes for more coarse GPS precision in the city. 
Therefore, while our work is a promising start for learning geometry from cross-view image pairs, there are still many challenges that need to be addressed.

\vspace{-2mm}
\section{Conclusion}
\vspace{-2mm}
In this paper, we propose a method, \ie Sat2Density, to learn a faithful 3D geometry representation of satellite scenes from satellite-ground image pairs through the satellite-to-ground-view synthesis task. Our approach tackles two critical issues, the infinity issue and the illumination difference issue, to make geometry learning possible. By leveraging the learned density, our model is capable of synthesizing spatial and temporal ground videos from satellite images even with only one-to-one satellite-ground image pairs for training. 
To the best of our knowledge, our method represents the first successful attempt to learn precise 3D geometry from satellite-ground image pairs, which significantly advances the recognition of satellite-ground tasks from a geometric perspective.

\paragraph{Acknowledgements.}
This work was supported by the National Nature Science Foundation of China under grant 62101390.

{\small
\bibliographystyle{ieee_fullname}
\bibliography{egbib}
}

\newpage

\section*{Appendix}
In the appendix, we first show our motivation and then present the details of the model architecture and training process of our Sat2Density model. After that, we describe the satellite and ground-view panorama camera models. Last, we give more discussion about our work for a better understanding of our paper and hope to advance viewing remote sensing and ground imagery from a geometric perspective. Code, pre-trained models, and more video results can be found on our project page.

\renewcommand\thesection{\Alph{section}}
\setcounter{section}{0}
\section{Motivation}
Sat2Density focuses on the geometric
nature of generating high-quality ground street views 
conditioned on satellite images learning from collections of
satellite-ground image pairs. The long-suffered issue from
the unknown 3D information is addressed by separating the
sky/non-sky regions with reasonable 3D density volumes
learned. We believe our new perspective on the longstanding yet challenging problem of satellite-ground novel view
synthesis would bring more insights for a wide range of 3D
vision tasks, including but not limited to (1) using satellite
images for autonomous driving with faithful 3D geometry,
(2) providing promising and novel solutions for visual localization with satellite images.

\section{Addition Implementation Details}
\subsection{DensityNet}
The DensityNet is taken from the generator of Pix2Pix~\cite{Pix2Pix}. Compared to vanilla Pix2Pix in PyTorch implementations from \href{https://github.com/junyanz/pytorch-CycleGAN-and-pix2pix}{pix2pix in PyTorch}, our generator replaces the activation function in the initial layer and downsample layers from ReLU to PReLU, sets the number of res-block to 6, and replaces ReLU with Tanh in the last layer. The final output of DensityNet is an explicit volume density $V_{\sigma}\in\mathbb{R}^{H\times W\times N}$, rather than predicting an image with resolution $H\times W\times 3$.

\subsection{Illumination Injection}
To inject the illumination, we first calculate the RGB histogram of the sky region in the ground image with 90 bins in each color channel. Following the way process style in GANcraft\cite{gancraft}, we use a style encoder to predict a style code, then use an MLP that is shared across all the style conditioning layers to convert the input style code to an intermediate illumination feature. The key difference is that the input of the style encoder is a histogram rather than an image. When inference, we could randomly select a histogram as the illumination input, at the same time, interpolation in the $z$ space between two histograms is also allowed. {\em The interpolation visualization video result can be seen on the project page.}

\subsection{RenderNet}
The RenderNet is a variation of Pix2Pix~\cite{Pix2Pix}. As shown in Figure~\ref{fig:RenderNet}, the key difference is that we inject the style feature in the last three Upsample blocks, which includes the illumination information of the groundtruth image during training, thereby mitigating the effects of illumination changes.

\begin{figure}[ht]
    \centering
    \resizebox{0.99\linewidth}{!}{
    \includegraphics{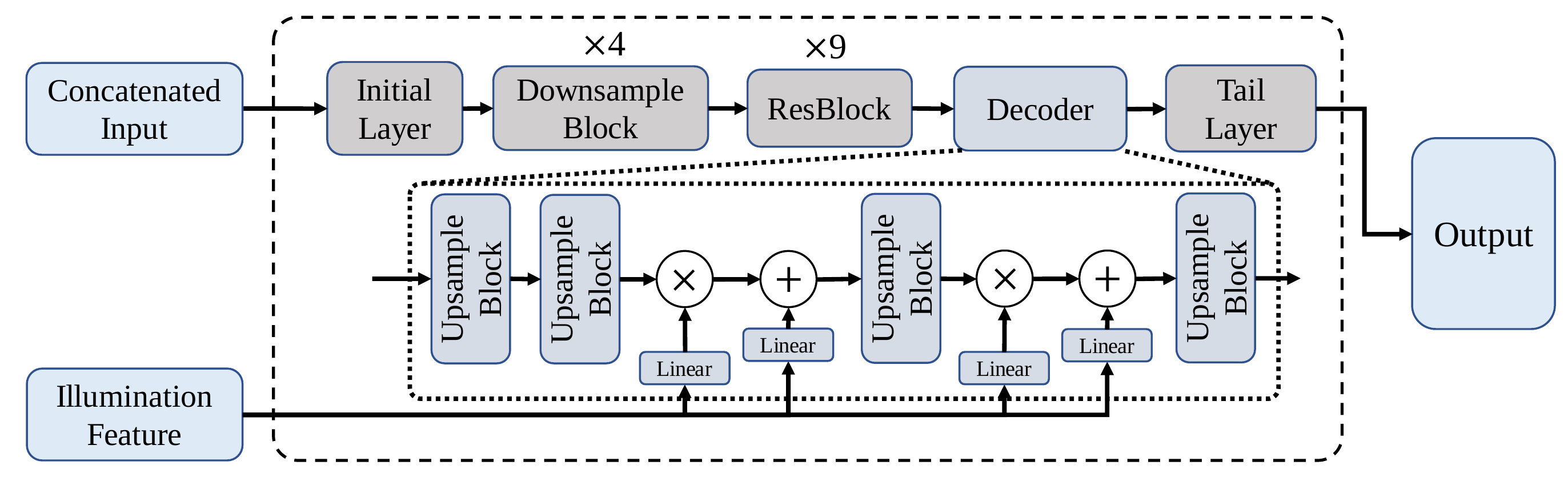}
    }
    \vspace{-1mm}
    \caption{The architecture of RenderNet. We inject the illumination feature in the decoder.}
    \label{fig:RenderNet}
    \vspace{-0.5 em}
\end{figure}

\subsection{Discriminator}

The discriminator we use is a multi-scale discriminator that differs from the vanilla multi-scale discriminator used in pix2pixHD~\cite{wang2018high}. While the vanilla discriminator operates on images of different scales, we use three discriminators: $D_1$, $D_2$, and $D_3$. $D_1$ works on panorama images, while $D_2$ and $D_3$ operate on perspective images obtained by randomly sampling from the input panorama using a perspective transformation, but at different scales. The two discriminators operate on perspective images because the distortion on the upper and lower bounds of the panorama is challenging for the convolution layer. Specifically, the field of view (FOV) of the sampled perspective images is 100. In our ablation study, all results use the same multi-scale discriminator. The input image size for $D_1$, $D_2$, and $D_3$ is $64 \times 256$, $64 \times 64$, and $32 \times 32$ respectively.

\subsection{Additional Training Details}
The weight for L1 loss, L2 loss, KL loss, feature matching loss, perceptual loss, $\mathcal{L}_{\rm snop}$ and GAN loss are 1, 10, 0.1, 10, 10, 1, 1 respectively when training. In volume rendering, we sample 100 points along each ray.

\section{Satellite and Panorama camera model}
Actually, there are no given camera instincts in the original CVUSA and CVACT datasets, which only contain image pairs collected from Google Earth in the same location by GPS, we follow the assumptions in Shi \etal~\cite{Shi}, which assumes that satellite images show the top of objects in an overhead view, which approximates parallel projection, while street-view panoramas capture scenes at ground level with a spherical equirectangular projection.

To describe a panoramic image with a 360-degree horizontal and 180-degree vertical field of view, we use the equirectangular projection and spherical coordinate system. To accomplish this, we assign the camera location as $\mathbf{o}$, and the width and height of the panorama image as $w$ and $h$, respectively. We use $x$ and $y$ as the pixel coordinates of the image pixel under consideration, and then we can use the following equations to determine the azimuthal and zenith angle $\theta$ and $\phi$, respectively:

$$\theta=\frac{2\pi x}{w}, \phi=\frac{\pi y}{h}$$

The equation allow us to determine the view direction $\mathbf{d}$ through any given image pixel.

We illustrate the orientation corresponding to the CVACT (align) dataset in Figure~\ref{fig:direction}, where the same color indicates the same direction.

\begin{figure}[ht]
    \vspace{-1em}
    \centering
    \subfigure[satellite]{
    \centering
    \includegraphics[width=0.25\linewidth, height = 0.25\linewidth]{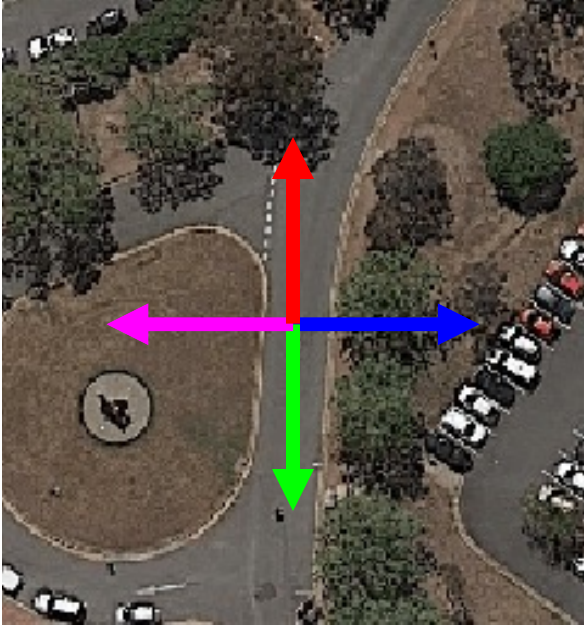}
    }
    \subfigure[panorama]{
    \centering
    \includegraphics[width=0.65\linewidth, height = 0.25\linewidth]{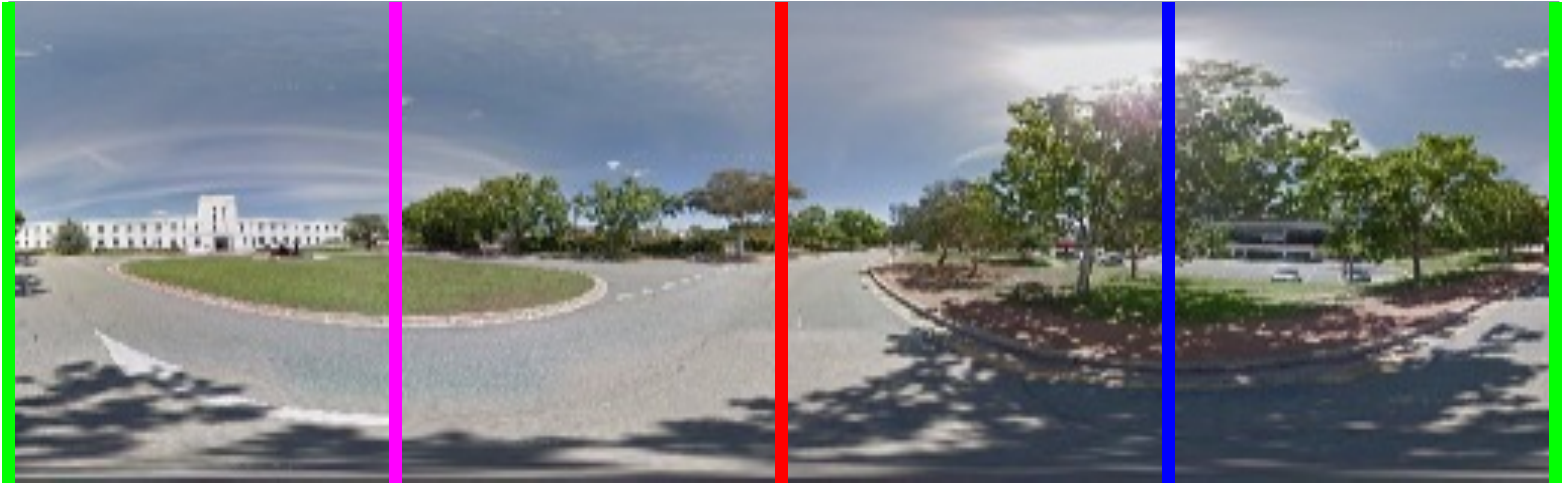}
    }
    \vspace{-1mm}
    \caption{Here is an example of an aligned satellite and ground panorama image pair from the training dataset. In the satellite image, the north direction is upward, while in the ground panorama image, the central column line represents the north direction. Both display the same red color. The central horizontal line in the panorama corresponds to the horizon.
    }
    \label{fig:direction}
    \vspace{-0.5 em}
\end{figure}

\section{Discussion}
\subsection{Urban scenes}
The nadir satellite image can not see the vertical surfaces of tall buildings. Except for the issue of unseen vertical surfaces, two representative cases of tall buildings and transient objects (\eg, cars) will challenge our method, though we believe the geometric perspective would facilitate the task for urban scenes.

\subsection{Infinite region}
 We assume that \textbf{objects beyond the top view coverage in street view images only include the sky}. It is hard to find out which object (e.g. tree) lies outside of the satellite scene, for we have no real 3D shape to find it. Nevertheless, our assumption has shown clear effects, as evident from the ablation study (check in the project page video).

 \subsection{Assumption on horizontal ground planes}
 Sat2Density has the horizontal assumption as we did not know the camera location and world coordinate system. Another assumption we used is that the ``world" is finite and limited by the satellite image. The used datasets follow these two assumptions and provide 1-to-1 paired data. Given by these facts, the movement of cameras is indeed on the ground plane with a constant height (\eg, 2m in our method and prior arts like \cite{Shi}). From the perspective of view synthesis, such assumptions should work well, but it will inherently lead to inaccurate 3D scene geometry when the whole ground region of a scene is sloped. To resolve this problem, further studies could be explored with some new problems: (1) how to estimate the slope from a single orthogonally-rectified satellite image, and (2) how to define a proper world coordinate system and place the ground-view camera(s) in the world.

\end{document}